\documentclass{article}
\usepackage[preprint,nonatbib]{neurips_2024} 

\usepackage[numbers]{natbib}
\usepackage[utf8]{inputenc} 
\usepackage[T1]{fontenc}    
\usepackage{hyperref}       
\usepackage{url}            
\usepackage{booktabs}       
\usepackage{amsfonts}       
\usepackage{nicefrac}       
\usepackage{microtype}      
\usepackage{xcolor}         
\usepackage{float}
\usepackage{amsmath}

\usepackage{enumitem}
\usepackage[caption=true]{subfig}
\usepackage{psfrag}
\usepackage{verbatim}
\usepackage{mathrsfs}
\usepackage{amssymb}%
\usepackage{pifont}%
\usepackage{multirow}
\usepackage{longtable}
\usepackage{listings}
\usepackage{graphicx}
\usepackage{amsmath}
\usepackage{mathtools}
\usepackage{amsthm}
\usepackage{xspace}
\usepackage{tcolorbox}
\usepackage{graphbox} 
\usepackage{colortbl}
\usepackage{wrapfig}

\definecolor{berkeleyblue}{HTML}{3B7EA1}
\definecolor{berkeleygold}{HTML}{FDB515}
\definecolor{main}{HTML}{4472C4}    
\definecolor{sub}{HTML}{EBF4FF}     

\newcommand{\DATASET}{\texttt{PersonaLLM}\xspace}
\newcommand{\BASELINEDIRECT}{PersonaLLM\xspace}

\title{Rediscovering the Latent Dimensions of Personality

with Large Language Models as Trait Descriptors}
\newcommand\blfootnote[1]{%
  \begingroup
  \renewcommand\thefootnote{}\footnote{#1}%
  \addtocounter{footnote}{-1}%
  \endgroup
}

\author{
Joseph Suh\blfootnote{Equal contribution}$^{*}$ \enspace\enspace Suhong Moon$^{*}$ \enspace\enspace Minwoo Kang$^{*}$ \enspace\enspace David M. Chan\\
\\
{University of California, Berkeley}\vspace{3mm}\\
    {{\small{\{josephsuh, suhong.moon, minwoo\_kang, davidchan\}@berkeley.edu}}}\vspace{-5mm}\\
}
\date{}

\begin{document}
\maketitle
\begin{abstract}

Assessing personality traits using large language models (LLMs) has emerged as an interesting and challenging area of research. While previous methods employ explicit questionnaires, often derived from the Big Five model of personality, we hypothesize that LLMs implicitly encode notions of personality when modeling next-token responses. To demonstrate this, we introduce a novel approach that uncovers latent personality dimensions in LLMs by applying singular value decomposition (SVD) to the log-probabilities of trait-descriptive adjectives. Our experiments show that LLMs ``rediscover'' core personality traits such as extraversion, agreeableness, conscientiousness, neuroticism, and openness without relying on direct questionnaire inputs, with the top-5 factors corresponding to Big Five traits explaining 74.3\% of the variance in the latent space. 
Moreover, we can use the derived principal components to assess personality along the Big Five dimensions, and achieve improvements in average personality prediction accuracy of up to 5\% over fine-tuned models, and up to 21\% over direct LLM-based scoring techniques.

\end{abstract}
\section{Introduction}
\label{sec:intro}

Over the past decades, researchers in personality psychology have investigated the viability of a general, descriptive model that yields systematic categorization of individual differences in personalities. 
These efforts culminated in one of the most widely-accepted taxonomies---the \textit{Big Five} model \citep{john2008paradigm, goldberg2013alternative} with extraversion, agreeableness, conscientiousness, neuroticism, and openness as five principal dimensions of personality traits---that has been consistently cross-validated in various studies with different human populations and assessment methods \citep{mccrae1992introduction, john1999big, costa1999five, gosling2003very}.
The core underlying principle behind the Big Five is the \textit{lexical hypothesis}, which asserts that the fundamental characteristics of personality are likely encoded in natural language, often as single words \citep{allport1937personality, goldberg1981language, raad1998lingua, john2008paradigm}.
The five factors were derived from analyses of terms that people frequently use to describe personalities of themselves and others, resulting in lists of trait descriptive adjectives (TDAs) \citep{norman1967estimating, goldberg1992development, goldberg2013alternative} that form the basis of psycholexical analyses and have influenced questionnaire-based assessments employed to-date, such as the Big Five Inventory (BFI) \citep{soto2017next} and NEO-FFI \citep{mccrae1992introduction, costa1999five}. 

More recently, advancements in large language models (LLMs) have sparked investigations of whether these models are capable of recognizing and expressing personality traits as humans \citep{personallm, amin2023will, serapio2023personality, peters2024large}, with potential applications in automatic text-based personality recognition \citep{mairesse2007using, kazameini2020personality, zhao2022deep, cao2024large, wen2024affective}.
As LLMs are trained on increasing amounts of text corpora, the expectation is that these models could correlate linguistic features reflected in samples of text to the personality dimensions of the authors \citep{perez2023discovering,anthology}.
Prior work explores direct prompting, \textit{i.e.} asking to directly predict the Big Five trait scales, or prompting to complete questionnaires \citep{personallm, serapio2023personality}.
While these work report that personality prediction may benefit from the use of LLMs, the dependence on direct prompts and questionnaires has potential risks of unreliable and prompt-sensitive results \citep{suhr2023challenging, gupta2024selfassessmenttestsunreliablemeasures, frisch2024llm}.

In this work, we first revisit the lexical hypothesis and the analyses that led to the discovery of the Big Five model: does LLMs' description of personality encode latent structures that align with the Big Five? 
We examine whether the same principle dimensions reported in human studies emerge when LLMs are prompted to describe personalities without imposing pre-defined taxonomies, thereby allowing natural, unrestricted depiction of personality traits.
Specifically, we condition language models so as to measure the likelihoods of each term in the Goldberg's 100-TDA \citep{goldberg1992development} in describing the author, over a set of synthetic personal stories.
These likelihoods, computed as log-probabilities, are analyzed using singular value decomposition (SVD) to uncover underlying personality factors.

The uncovered factors are then used to predict the polarity of Big Five traits of the authors of unseen stories.
Overall, our results demonstrate not only that LLMs encode a latent structure similar to the Big Five model, with the five principal factors explaining 74.3\% of total variance, but also our approach to personality trait predictions outperform previous methods, by up to 5\% over models fine-tuned on Big Five trait classification and by up to 21\% over direct LLM-scoring approaches.
\begin{figure*}[t]
    \centering
    \captionsetup{font=small}
    \includegraphics[width=0.99\linewidth]{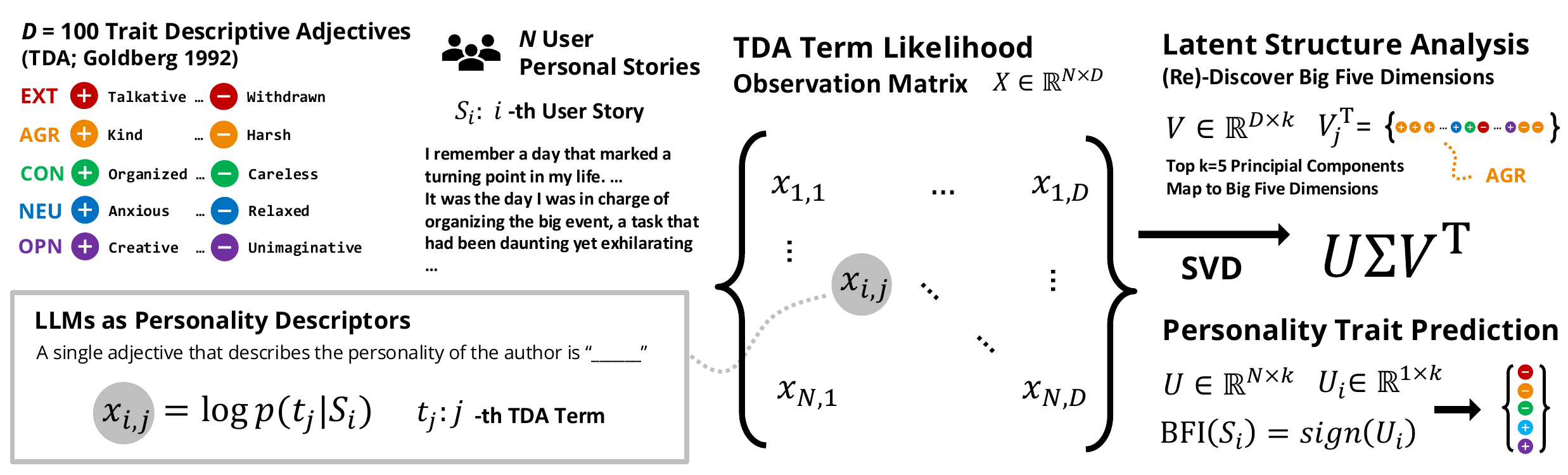}
    \vspace{-5pt}
    \caption{
    Overview of our approach.
    Given a set of personal stories and a list of TDAs, a language model computes the likelihood of each trait term describing the author of the story, through which we construct an observation matrix of such log-probabilities.
    Singular Value Decomposition (SVD) is then applied to this matrix, which yields (1) the loading matrix $V$ that captures the latent dimension structure and (2) the factor matrix $U$ that explains each story author's personality spectrum in the projected low-dimensional space.
    Results are compared against findings from psychometric literature and binary Big Five labels of authors, respectively.
    }
    \label{fig:worfklow}
    \vspace{-5pt}
\end{figure*}

\begin{wrapfigure}{r}{0.45\textwidth}
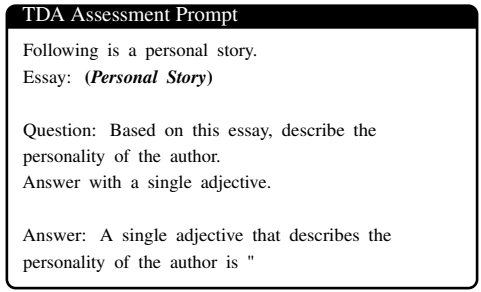

    \vspace{-10pt}
    \captionsetup{font=small}
    \begin{tcolorbox}[width=1.0\linewidth, halign=left, colframe=black, colback=white, boxsep=0.01mm, arc=1.0mm, left=2mm, right=2mm, boxrule=1pt, title=\fontsize{8pt}{10pt}\selectfont{ TDA Assessment Prompt},]
    \fontsize{7pt}{10pt}\selectfont{
    Following is a personal story.\\
    Essay: \textbf{(\textit{Personal Story})}
    \newline\newline
    Question: Based on this essay, describe the personality of the author. \\
    Answer with a single adjective.
    \newline\newline
    Answer: A single adjective that describes the personality of the author is "
    }
    \end{tcolorbox}
    \caption{A question-answer format prompt that asks for a single adjective describing the personality of the author of the given personal story. A quotation mark is adopted at the prompt suffix to guide the model to complete the sentence with an adjective. For each adjective in the TDA, we measure the log-probability of the model completing the prompt with the given adjective.
    }
    \label{box:question_example_main}
\end{wrapfigure}

\section{Using Language Models as Personality Descriptors}
\label{sec:method}

In this work, we hypothesize that LLMs encode personality traits implicitly within their latent spaces. More explicitly, we argue that we can analyze and extract personality traits present in personal stories by carefully querying the LLM and analyzing the log-probabilities of trait-descriptive adjectives (TDAs) within the next token prediction. Here we ask: is it possible to recover traditional categories of personality from the principal components of the resulting token likelihood matrix?

We present the primary steps of our approach in Figure \ref{fig:worfklow}.
First, we measure the likelihood of a set of personality trait terms describing the story reflecting each individual author, \textit{i.e.} calculate the log-probabilities of the 100 adjectives in the TDA curated in \cite{goldberg1992development}.
In doing so, we prompt LLMs with a personal story as illustrated in Figure~\ref{box:question_example_main}.
Then, we build an observation matrix where each row contains the log-probabilities of TDA for each user.
These log-probabilities can capture the usage frequency of adjectives varies in real life, based on the input prompt.
For instance, adjectives associated with extraversion are more commonly used than those related to conscientiousness, leading to differences in latent factor levels \citep{goldberg1981language}.
Therefore, this subtle difference is captured in log-probabilites, allowing the method to recover the five principal axes corresponding to each trait in the Big Five model.
Further details on the measurement process can be found in Appendix \ref{asec:logprob_measure}.

After building the observation matrix, we perform singular value decomposition (SVD) on the matrix to identify the top-5 factors. 
Based on the SVD results, we analyze both the loading matrix and the factor matrix (Section~\ref{sec:experiment}). 
We then report the classification results derived from the factor matrix and compare these with the findings from the previous psychometric literature~\citep{john1999big,goldberg1981language,goldberg1992development,goldberg2013alternative}.

\subsection{Re-Discovering the Big Five Structure}
We start with measuring the log-probabilities of the trait terms in describing each of the users and construct the observation matrix $X \in  \mathbb{R}^{N\times D}$, where $N$ is the number of users and $D$ is the number of trait adjectives in Figure~\ref{fig:worfklow}.
We observe that the log-probabilities are biased by the frequency of usage of trait adjectives, as shown in Figure~\ref{fig:logprob_mean_std}, and accordingly apply zero-centering.
SVD is applied to the matrix $X$, resulting in $\tilde{X} = U\Sigma V^\top$---in this decomposition, $U \in \mathbb{R}^{N \times k}$ is the \textbf{factor matrix} that represents the latent features of personality traits of each story author, $\Sigma \in \mathbb{R}^{k \times k}$ contains the singular values, and $V \in \mathbb{R}^{D \times k}$ is the \textbf{loading matrix} that 
defines how each trait adjective relates to the latent factors. 
We compare the factor matrix with factor loading identified in \cite{goldberg1992development} to examine whether the principal components uncovered by the factor matrix resembles the Big Five dimensions.

\subsection{Personality Trait Prediction} 
Each element of the factor matrix $U_{ij}$ represents the extent of the \textit{i}-th story's association with the \textit{j}-th latent factor.
Specifically, the sign of the element predicts the binary personality trait (e.g., `extraverted' or `introverted') of the story author, and for evaluation, we compare these polarities against the actual personality trait labels.
Since our approach is based on an \textit{unsupervised} method, the full procedure does not require explicit labels of personality traits.
When labels are available, we can further enhance our analysis by incorporating supervised learning.
Specifically, we choose Lasso regression using the log-probability features to predict the personality traits since Lasso regression performs feature selection by identifying the most informative trait adjectives.
\section{Experiments}
\label{sec:experiment}

We benchmark our methods against baselines on \DATASET dataset~\citep{personallm}. \DATASET dataset is generated by \texttt{gpt-4-0613}~\citep{gpt4}, 
where \texttt{gpt-4-0613} is prompted with Big Five traits and asked to generate stories consistent with the prescribed personality.
More details about the dataset are in Appendix~\ref{asec:dataset}.
We consider a suite of LLMs including Meta Llama 3 and Llama 3.1 model families \citep{llama3} and Mixtral 8x22B \citep{Jiang2024Mixtral, mixtral-8x22b}.
We present the analysis results that uncover the Big Five structure in Section~\ref{subsec:big_five_structure} and compare trait prediction accuracy with baselines in Section~\ref{subsec:predicting_personality_trait}.

\subsection{The Big Five Structure}
\label{subsec:big_five_structure}
The sum of the top-5 squared singular values from SVD accounts for 74.3\% of the total variance in the log-probability measurements, indicating that it sufficiently captures the variance.
The detailed statistics of singular values are provided in Appendix \ref{asec:sub:svd_sing_val}.
We compare elements of principal components with factor loadings from the literature \citep{goldberg1992development}.
Specifically, identifying which adjectives have large elements in each principal component establishes the correspondence between principal components and personality factors. For example, since adjectives that have high loading in the extraversion factor (e.g., `energetic', `bashful', `reserved') also have large elements in the first principal component, the first principal component relates to the extraversion trait.
We identify the one-to-one correspondence between principal components and Big Five traits; the top-5 principal components correspond with extraversion, openness, agreeableness, neuroticism, and conscientiousness traits, respectively.
Detailed steps for finding the correspondence are given in Appendix \ref{asec:sub:factor_loading}.

\begin{table}[t]
    \small
    \captionsetup{font=small}
    \centering
    \caption{
    Test accuracy of binary personality assessment using various approaches. 
    The first two rows represent baseline methods—\BASELINEDIRECT and fine-tuning DeBERTaV3 \cite{he2021debertav3}. 
    For \BASELINEDIRECT, we use the numbers reported in~\cite{personallm}.
    Predictions obtained using SVD are comparable to that of the baseline methods, while regression performed on a set of trait adjectives surpasses other methods. 
    `Instruct' denotes instruction fine-tuned, chat models; otherwise, if unspecified, the pre-trained model is used. 
    Bold-faced numbers indicate the highest accuracy for each of the Big Five traits.
    }
    \vspace{5pt}
    \resizebox{1.0\linewidth}{!}{%
        \begin{tabular}{llcccccc}
        \toprule
        \multirow{2.5}{*}{Method} & \multirow{2.5}{*}{Model} & \multicolumn{6}{c}{Big Five Trait Prediction Accuracy}\\
         \cmidrule{3-8}
         &  & Extraversion & Agreeableness & Conscientiousness & Neuroticism & Openness & Avg.\\
        \midrule
        \BASELINEDIRECT  & \texttt{gpt-4-0613} & 0.97 & 0.69 & 0.69 & 0.56 & 0.59 & 0.698\\
        \midrule
        Encoder FT & \texttt{DeBERTaV3} & \textbf{1.000} & 0.833 & 0.786 & 0.833 & 0.881 & 0.867  \\
        \midrule
        \multirow{9}{*}{SVD} 
        & \texttt{Llama-3.1-70B } & 0.905 & 0.786 & 0.726 & 0.774 & 0.762 & 0.791 \\
        & \texttt{Llama-3.1-70B-Instruct} & 0.917 & 0.798 & 0.536 & 0.607 & 0.679 & 0.707 \\
        & \texttt{Llama-3-70B }   & 0.988 & 0.798 & 0.667 & 0.655 & 0.810 & 0.783 \\
        & \texttt{Llama-3-70B-Instruct}   & 0.905 & 0.845 & 0.786 & 0.607 & 0.821 & 0.793 \\
        \cmidrule{2-8}
        & \texttt{Llama-3.1-8B }  & 0.857 & 0.702 & 0.821 & 0.690 & 0.667 & 0.748 \\
        & \texttt{Llama-3.1-8B-Instruct}  & 0.702 & 0.810 & 0.560 & 0.738 & 0.524 & 0.667 \\
        & \texttt{Llama-3-8B }    & 0.869 & 0.750 & 0.786 & 0.726 & 0.619 & 0.750 \\
        & \texttt{Llama-3-8B-Instruct}    & 0.905 & 0.679 & 0.583 & 0.738 & 0.548 & 0.690 \\
        \cmidrule{2-8}
        & \texttt{Mixtral-8x22B } & 0.917 & 0.798 & 0.536 & 0.607 & 0.679 & 0.707 \\

        \midrule
        \multirow{9}{*}{Lasso} 
        & \texttt{Llama-3.1-70B } & 0.964 & \textbf{0.964} & 0.881 & \textbf{0.893} & 0.857 & \textbf{0.912} \\
        & \texttt{Llama-3.1-70B-Instruct} & 0.988 & 0.845 & \textbf{0.857} & 0.857 & \textbf{0.905} & 0.890 \\
        & \texttt{Llama-3-70B }   & \textbf{1.000} & 0.893 & 0.845 & 0.881 & \textbf{0.905} & 0.905 \\
        & \texttt{Llama-3-70B-Instruct}   & \textbf{1.000} & 0.833 & 0.821 & 0.869 & 0.845 & 0.874 \\
        \cmidrule{2-8}
        & \texttt{Llama-3.1-8B }  & 0.964 & 0.869 & 0.821 & \textbf{0.893} & 0.881 & 0.886 \\
        & \texttt{Llama-3.1-8B-Instruct}  & 0.964 & 0.857 & 0.786 & 0.845 & 0.845 & 0.860 \\
        & \texttt{Llama-3-8B }    & 0.988 & 0.857 & 0.798 & \textbf{0.893} & 0.893 & 0.886 \\
        & \texttt{Llama-3-8B-Instruct}    & 0.988 & 0.857 & 0.786 & 0.833 & 0.833 & 0.860 \\
        \cmidrule{2-8}
        & \texttt{Mixtral-8x22B } & 0.988 & 0.845 & \textbf{0.857} & 0.857 & \textbf{0.905} & 0.890 \\
        
        \bottomrule
        \end{tabular}
    }
    \label{tab:ours_accuracy}
\end{table}

\subsection{Personality Trait Prediction}
\label{subsec:predicting_personality_trait}
$U$ matrix of SVD can be interpreted as the personality trait scales of each story author.
More specifically, the element of $U$, $U_{ij}$, represents the scale of the $j$-th Big Five trait of the $i$-th user.
By taking the signs of the predicted scales and comparing them with binary personality labels, we measure the accuracy of personality assessment.
As a baseline, we employ (1) fine-tuning DeBERTaV3 \citep{he2021debertav3} for the Big Five binary label classification task, and (2) prompting LLMs to evaluate the personality score directly.
The details of each baseline method are presented in Appendix \ref{asec:baseline}.

The personality prediction accuracy resulting from SVD is comparable to that of the baseline methods.
Additionally, we perform Lasso regression using the log-probabilities of trait adjectives as features and binary personality labels as targets. 
This regression method outperforms other approaches across all personality traits.
\vspace{-3pt}
\section{Further Directions}
\label{sec:impact}
\vspace{-5pt}

While the results in latent dimension analysis and personality assessment are promising, several psychometric aspects deserve further exploration.

\textbf{Convergent Validity:} Convergent validity assesses the extent to which different measures that are theoretically related are also experimentally related. To examine this aspect, we need to verify whether a set of widely accepted personality tests, as well as self-perception and external perception tests, consistently reveal a latent structure in personality assessment that aligns with human perception.

\textbf{Role of Acquaintance:} The level of acquaintance is known to influence the accuracy of personality descriptions. Studies indicate that acquaintance between a person and a perceiver generally enhances the assessment accuracy \cite{funder1988friends, lee2017acquaintanceship}. We may simulate acquaintance with persona steering methods \cite{anthology} and perform personality assessments at varying levels of acquaintance to investigate the existence of the same correlation in LLMs.

\textbf{Test-retest Reliability:} Test-retest reliability evaluates the consistency of test results by administering the same test to the same group at different points in time. It could be assessed through behavioral change experiments where language models are in conversational settings. By comparing personality assessments at the beginning and end of the interaction, we may investigate the test-retest reliability.
\vspace{-3pt}
\section{Conclusion}
\label{sec:discussion}
\vspace{-5pt}
This work introduces a method to probe the latent dimensions of personality using LLMs as trait descriptors. By analyzing key characteristics of principal components, we observe that the personality perception by LLMs encodes a latent structure closely resembling that of human perception. Our method also serves as a robust tool for personality assessment that matches or surpasses the accuracy of previous LLM-based approaches. Building on one of the most intrinsic aspects of human--personality--this framework holds the potential to explore a vast array of real-world values, including ethical and cultural norms, and social behaviors.

\clearpage
\bibliographystyle{plainnat}
\bibliography{arxiv_reference}

\begin{thebibliography}{36}
\providecommand{\natexlab}[1]{#1}
\providecommand{\url}[1]{\texttt{#1}}
\expandafter\ifx\csname urlstyle\endcsname\relax
  \providecommand{\doi}[1]{doi: #1}\else
  \providecommand{\doi}{doi: \begingroup \urlstyle{rm}\Url}\fi

\bibitem[Achiam et~al.(2023)Achiam, Adler, Agarwal, Ahmad, Akkaya, Aleman, Almeida, Altenschmidt, Altman, Anadkat, et~al.]{gpt4}
Josh Achiam, Steven Adler, Sandhini Agarwal, Lama Ahmad, Ilge Akkaya, Florencia~Leoni Aleman, Diogo Almeida, Janko Altenschmidt, Sam Altman, Shyamal Anadkat, et~al.
\newblock Gpt-4 technical report.
\newblock \emph{arXiv preprint arXiv:2303.08774}, 2023.

\bibitem[AI@Meta(2024)]{llama3}
AI@Meta.
\newblock Llama 3 model card.
\newblock 2024.
\newblock URL \url{https://github.com/meta-llama/llama3/blob/main/MODEL_CARD.md}.

\bibitem[Allport(1937)]{allport1937personality}
Gordon~Willard Allport.
\newblock Personality: A psychological interpretation.
\newblock 1937.

\bibitem[Amin et~al.(2023)Amin, Cambria, and Schuller]{amin2023will}
Mostafa~M Amin, Erik Cambria, and Bj{\"o}rn~W Schuller.
\newblock Will affective computing emerge from foundation models and general artificial intelligence? a first evaluation of chatgpt.
\newblock \emph{IEEE Intelligent Systems}, 38\penalty0 (2):\penalty0 15--23, 2023.

\bibitem[Cao and Kosinski(2024)]{cao2024large}
Xubo Cao and Michal Kosinski.
\newblock Large language models know how the personality of public figures is perceived by the general public.
\newblock \emph{Scientific Reports}, 14\penalty0 (1):\penalty0 6735, 2024.

\bibitem[Costa and McCrae(1999)]{costa1999five}
PT~Costa and RR~McCrae.
\newblock A five-factor theory of personality.
\newblock \emph{Handbook of personality: Theory and research}, 2\penalty0 (01):\penalty0 1999, 1999.

\bibitem[Frisch and Giulianelli(2024)]{frisch2024llm}
Ivar Frisch and Mario Giulianelli.
\newblock Llm agents in interaction: Measuring personality consistency and linguistic alignment in interacting populations of large language models.
\newblock \emph{arXiv preprint arXiv:2402.02896}, 2024.

\bibitem[Funder and Colvin(1988)]{funder1988friends}
David~C Funder and C~Randall Colvin.
\newblock Friends and strangers: acquaintanceship, agreement, and the accuracy of personality judgment.
\newblock \emph{Journal of personality and social psychology}, 55\penalty0 (1):\penalty0 149, 1988.

\bibitem[Goldberg(1981)]{goldberg1981language}
Lewis~R Goldberg.
\newblock Language and individual differences: The search for universals in personality lexicons.
\newblock \emph{Review of Personality and Social Psychology/Sage}, 1981.

\bibitem[Goldberg(1992)]{goldberg1992development}
Lewis~R Goldberg.
\newblock The development of markers for the big-five factor structure.
\newblock \emph{Psychological assessment}, 4\penalty0 (1):\penalty0 26, 1992.

\bibitem[Goldberg(2013)]{goldberg2013alternative}
Lewis~R Goldberg.
\newblock An alternative “description of personality”: The big-five factor structure.
\newblock In \emph{Personality and Personality Disorders}, pages 34--47. Routledge, 2013.

\bibitem[Gosling et~al.(2003)Gosling, Rentfrow, and Swann~Jr]{gosling2003very}
Samuel~D Gosling, Peter~J Rentfrow, and William~B Swann~Jr.
\newblock A very brief measure of the big-five personality domains.
\newblock \emph{Journal of Research in personality}, 37\penalty0 (6):\penalty0 504--528, 2003.

\bibitem[Gupta et~al.(2024)Gupta, Song, and Anumanchipalli]{gupta2024selfassessmenttestsunreliablemeasures}
Akshat Gupta, Xiaoyang Song, and Gopala Anumanchipalli.
\newblock Self-assessment tests are unreliable measures of llm personality, 2024.
\newblock URL \url{https://arxiv.org/abs/2309.08163}.

\bibitem[He et~al.(2021)He, Gao, and Chen]{he2021debertav3}
Pengcheng He, Jianfeng Gao, and Weizhu Chen.
\newblock Debertav3: Improving deberta using electra-style pre-training with gradient-disentangled embedding sharing. corr abs/2111.09543 (2021).
\newblock \emph{arXiv preprint arXiv:2111.09543}, 2021.

\bibitem[Jiang et~al.(2024)Jiang, Sablayrolles, Roux, Mensch, Savary, Bamford, Chaplot, de~Las~Casas, Hanna, Bressand, Lengyel, Bour, Lample, Lavaud, Saulnier, Lachaux, Stock, Subramanian, Yang, Antoniak, Scao, Gervet, Lavril, Wang, Lacroix, and Sayed]{Jiang2024Mixtral}
Albert~Q. Jiang, Alexandre Sablayrolles, Antoine Roux, Arthur Mensch, Blanche Savary, Chris Bamford, Devendra~Singh Chaplot, Diego de~Las~Casas, Emma~Bou Hanna, Florian Bressand, Gianna Lengyel, Guillaume Bour, Guillaume Lample, L'elio~Renard Lavaud, Lucile Saulnier, Marie-Anne Lachaux, Pierre Stock, Sandeep Subramanian, Sophia Yang, Szymon Antoniak, Teven~Le Scao, Th{\'e}ophile Gervet, Thibaut Lavril, Thomas Wang, Timoth{\'e}e Lacroix, and William~El Sayed.
\newblock Mixtral of experts.
\newblock \emph{ArXiv}, abs/2401.04088, 2024.
\newblock URL \url{https://api.semanticscholar.org/CorpusID:266844877}.

\bibitem[Jiang et~al.(2023)Jiang, Zhang, Cao, Kabbara, and Roy]{personallm}
Hang Jiang, Xiajie Zhang, Xubo Cao, Jad Kabbara, and Deb Roy.
\newblock Personallm: Investigating the ability of gpt-3.5 to express personality traits and gender differences.
\newblock \emph{arXiv preprint arXiv:2305.02547}, 2023.

\bibitem[John et~al.(1999)John, Srivastava, et~al.]{john1999big}
Oliver~P John, Sanjay Srivastava, et~al.
\newblock The big-five trait taxonomy: History, measurement, and theoretical perspectives.
\newblock 1999.

\bibitem[John et~al.(2008)John, Naumann, and Soto]{john2008paradigm}
Oliver~P John, Laura~P Naumann, and Christopher~J Soto.
\newblock Paradigm shift to the integrative big five trait taxonomy.
\newblock \emph{Handbook of personality: Theory and research}, 3\penalty0 (2):\penalty0 114--158, 2008.

\bibitem[Kazameini et~al.(2020)Kazameini, Fatehi, Mehta, Eetemadi, and Cambria]{kazameini2020personality}
Amirmohammad Kazameini, Samin Fatehi, Yash Mehta, Sauleh Eetemadi, and Erik Cambria.
\newblock Personality trait detection using bagged svm over bert word embedding ensembles.
\newblock \emph{arXiv preprint arXiv:2010.01309}, 2020.

\bibitem[Kwon et~al.(2023)Kwon, Li, Zhuang, Sheng, Zheng, Yu, Gonzalez, Zhang, and Stoica]{vllm}
Woosuk Kwon, Zhuohan Li, Siyuan Zhuang, Ying Sheng, Lianmin Zheng, Cody~Hao Yu, Joseph Gonzalez, Hao Zhang, and Ion Stoica.
\newblock Efficient memory management for large language model serving with pagedattention.
\newblock In \emph{Proceedings of the 29th Symposium on Operating Systems Principles}, pages 611--626, 2023.

\bibitem[Lee and Ashton(2017)]{lee2017acquaintanceship}
Kibeom Lee and Michael~C Ashton.
\newblock Acquaintanceship and self/observer agreement in personality judgment.
\newblock \emph{Journal of Research in Personality}, 70:\penalty0 1--5, 2017.

\bibitem[Loshchilov et~al.(2017)Loshchilov, Hutter, et~al.]{loshchilov2017fixing}
Ilya Loshchilov, Frank Hutter, et~al.
\newblock Fixing weight decay regularization in adam.
\newblock \emph{arXiv preprint arXiv:1711.05101}, 5, 2017.

\bibitem[Mairesse et~al.(2007)Mairesse, Walker, Mehl, and Moore]{mairesse2007using}
Fran{\c{c}}ois Mairesse, Marilyn~A Walker, Matthias~R Mehl, and Roger~K Moore.
\newblock Using linguistic cues for the automatic recognition of personality in conversation and text.
\newblock \emph{Journal of artificial intelligence research}, 30:\penalty0 457--500, 2007.

\bibitem[McCrae and John(1992)]{mccrae1992introduction}
Robert~R McCrae and Oliver~P John.
\newblock An introduction to the five-factor model and its applications.
\newblock \emph{Journal of personality}, 60\penalty0 (2):\penalty0 175--215, 1992.

\bibitem[MistralAI(2024)]{mixtral-8x22b}
MistralAI.
\newblock Mixtral-8x22b, 2024.
\newblock URL \url{https://mistral.ai/news/mixtral-8x22b/}.

\bibitem[Moon et~al.(2024)Moon, Abdulhai, Kang, Suh, Soedarmadji, Behar, and Chan]{anthology}
Suhong Moon, Marwa Abdulhai, Minwoo Kang, Joseph Suh, Widyadewi Soedarmadji, Eran~Kohen Behar, and David~M Chan.
\newblock Virtual personas for language models via an anthology of backstories.
\newblock \emph{arXiv preprint arXiv:2407.06576}, 2024.

\bibitem[Norman(1967)]{norman1967estimating}
Warren~T Norman.
\newblock On estimating psychological relationships: Social desirability and self-report.
\newblock \emph{Psychological Bulletin}, 67\penalty0 (4):\penalty0 273, 1967.

\bibitem[Perez et~al.(2023)]{perez2023discovering}
Ethan Perez et~al.
\newblock Discovering language model behaviors with model-written evaluations. arxiv, dec. 2022, 2023.

\bibitem[Peters and Matz(2024)]{peters2024large}
Heinrich Peters and Sandra Matz.
\newblock Large language models can infer psychological dispositions of social media users.
\newblock \emph{PNAS Nexus}, page pgae231, 2024.

\bibitem[Raad et~al.(1998)Raad, Perugini, Hreb{\'\i}ckov{\'a}, and Szarota]{raad1998lingua}
Boele~De Raad, Marco Perugini, Martina Hreb{\'\i}ckov{\'a}, and Piotr Szarota.
\newblock Lingua franca of personality: Taxonomies and structures based on the psycholexical approach.
\newblock \emph{Journal of Cross-Cultural Psychology}, 29\penalty0 (1):\penalty0 212--232, 1998.

\bibitem[Saucier(1997)]{saucier1997effects}
Gerard Saucier.
\newblock Effects of variable selection on the factor structure of person descriptors.
\newblock \emph{Journal of personality and social psychology}, 73\penalty0 (6):\penalty0 1296, 1997.

\bibitem[Serapio-Garc{\'\i}a et~al.(2023)Serapio-Garc{\'\i}a, Safdari, Crepy, Sun, Fitz, Romero, Abdulhai, Faust, and Matari{\'c}]{serapio2023personality}
Greg Serapio-Garc{\'\i}a, Mustafa Safdari, Cl{\'e}ment Crepy, Luning Sun, Stephen Fitz, Peter Romero, Marwa Abdulhai, Aleksandra Faust, and Maja Matari{\'c}.
\newblock Personality traits in large language models.
\newblock \emph{arXiv preprint arXiv:2307.00184}, 2023.

\bibitem[Soto and John(2017)]{soto2017next}
Christopher~J Soto and Oliver~P John.
\newblock The next big five inventory (bfi-2): Developing and assessing a hierarchical model with 15 facets to enhance bandwidth, fidelity, and predictive power.
\newblock \emph{Journal of personality and social psychology}, 113\penalty0 (1):\penalty0 117, 2017.

\bibitem[S{\"u}hr et~al.(2023)S{\"u}hr, Dorner, Samadi, and Kelava]{suhr2023challenging}
Tom S{\"u}hr, Florian~E Dorner, Samira Samadi, and Augustin Kelava.
\newblock Challenging the validity of personality tests for large language models.
\newblock \emph{arXiv e-prints}, pages arXiv--2311, 2023.

\bibitem[Wen et~al.(2024)Wen, Cao, Yang, Yang, and Liu]{wen2024affective}
Zhiyuan Wen, Jiannong Cao, Yu~Yang, Ruosong Yang, and Shuaiqi Liu.
\newblock Affective-nli: Towards accurate and interpretable personality recognition in conversation.
\newblock In \emph{2024 IEEE International Conference on Pervasive Computing and Communications (PerCom)}, pages 184--193. IEEE, 2024.

\bibitem[Zhao et~al.(2022)Zhao, Tang, and Zhang]{zhao2022deep}
Xiaoming Zhao, Zhiwei Tang, and Shiqing Zhang.
\newblock Deep personality trait recognition: a survey.
\newblock \emph{Frontiers in Psychology}, 13:\penalty0 839619, 2022.

\end{thebibliography}
\newpage
\appendix
\section{Log-probability Measurement Details}
\label{asec:logprob_measure}
\subsection{Why are TDA Log-probabilities Meaningful?}
\label{asec:sub:logprob_distribution}
\begin{figure*}[ht]
    \centering
    \captionsetup{font=small}
    \rotatebox{270}{\includegraphics[width=1.6\linewidth]{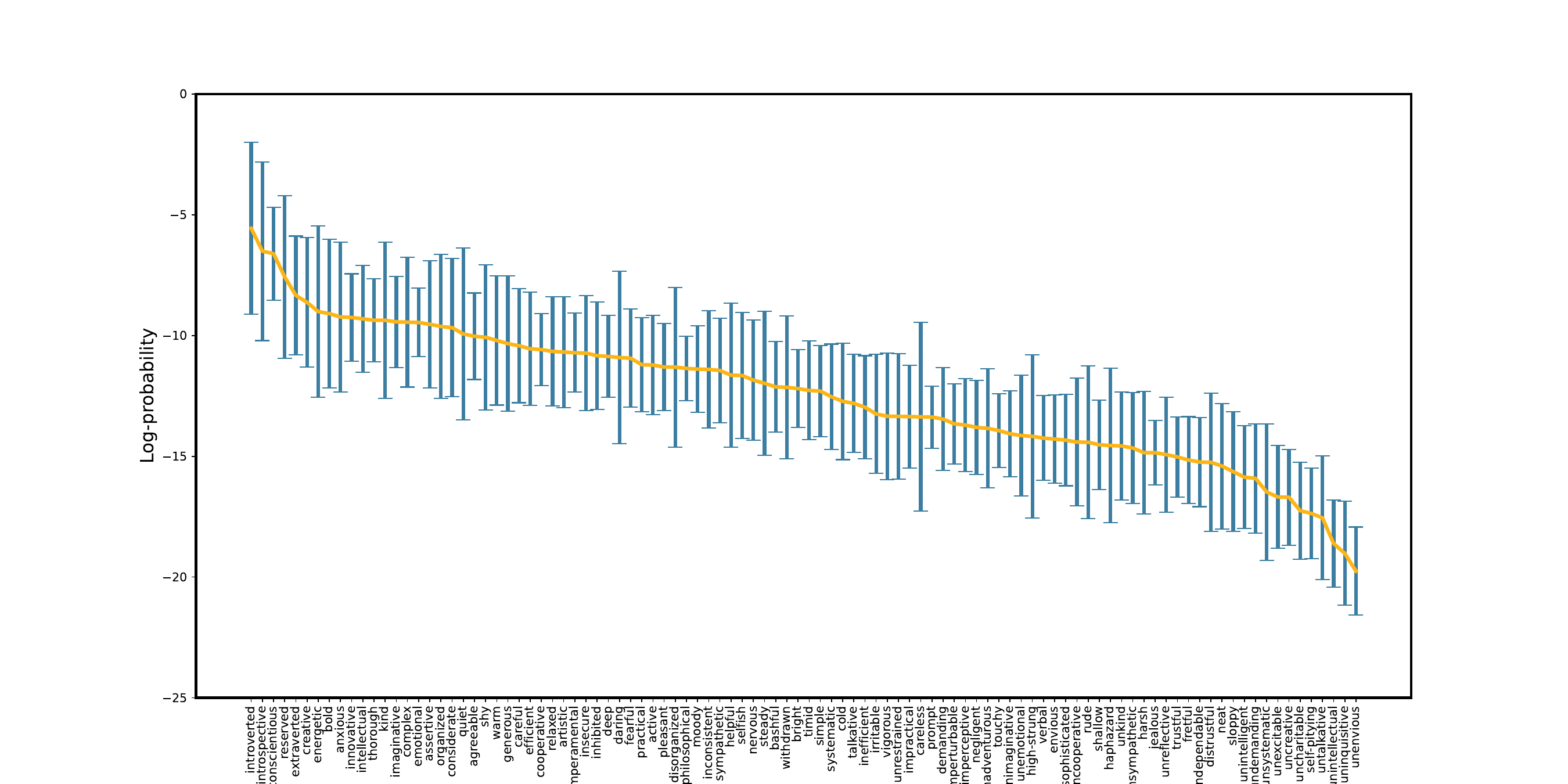}}
    \vspace{-50pt}
    \caption{
    Mean (gold) and standard deviation (blue) of log-probabilities for 100 trait adjectives, sorted by the mean. Log-probabilities are measured from the \texttt{PersonaLLM} dataset (Appendix \ref{asec:dataset}) with pretrained \texttt{Llama-3.1-70B}, decoding temperature $T=1$.
    }
    \label{fig:logprob_mean_std}
\end{figure*}

\begin{figure*}[ht]
    \centering
    \captionsetup{font=small}
    \includegraphics[width=1.0\linewidth]{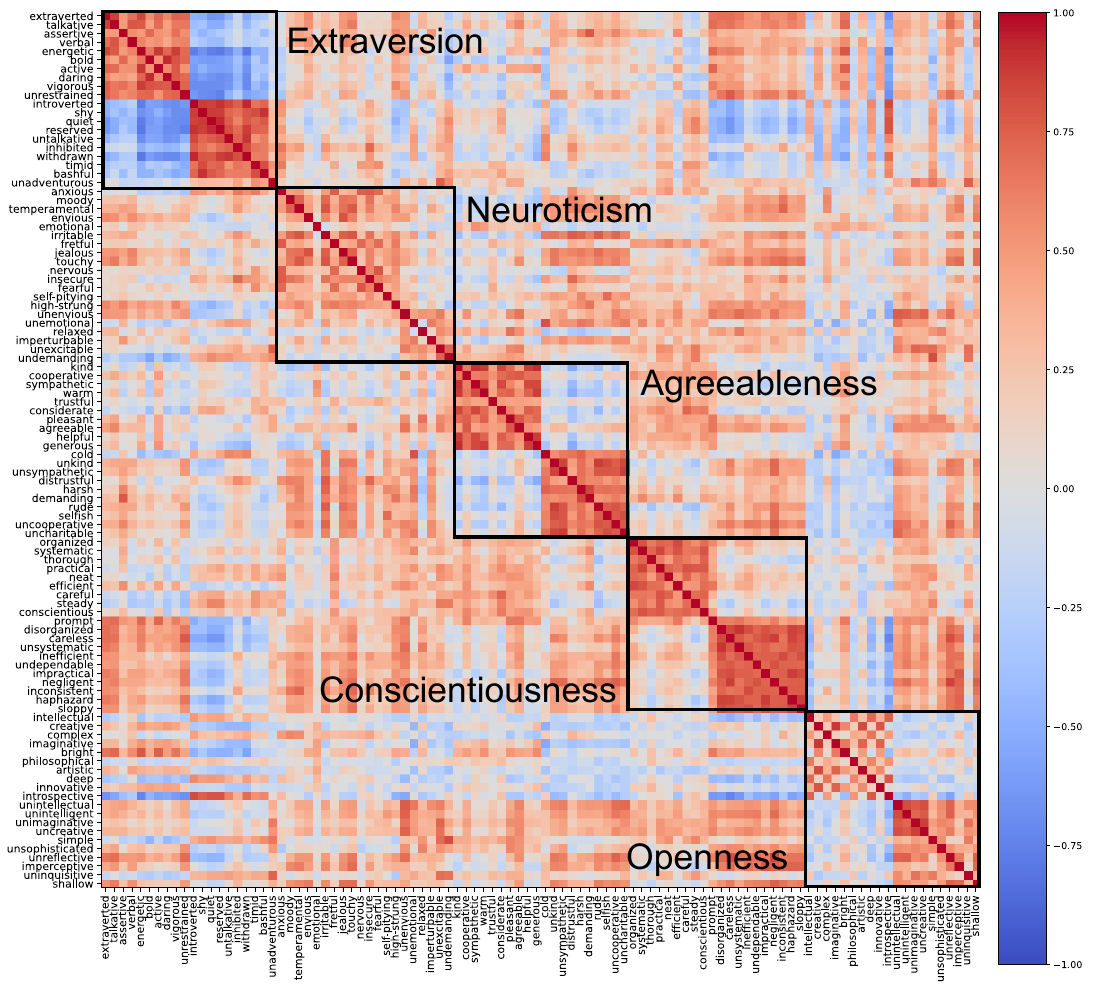}

    \vspace{-5pt}
    \caption{
    Visualization of correlation between trait adjectives. Datasets, models, and decoding parameters are identical to those of Figure \ref{fig:logprob_mean_std}. Five boxes with black edges indicate personality traits that adjectives belong to, drawn for visual aid. Trait adjectives that share Big Five trait show strong correlation, either positive or negative. We note that correlations between adjectives of different Big Five trait also show moderate level of correlation (e.g., `introverted' and `introspective'). This may imply that an adjective is related to several latent factors (instead of a single latent factdor) or that Big Five personality traits are not orthogonal.
    }
    \label{fig:corr_matrix}
\end{figure*}

We argue that log-probabilities of personality trait descriptive adjectives computed by LLMs are indicative measures of how accurately those adjectives describe a person.
While these log-probabilities encode likelihood that the model assigns to describing the person with a specific adjective, the prediction of such likelihoods are influenced not only by the explicit or implicit references to behavioral characteristics reflected in the story, but also by the patterns of general language use. 
Figure \ref{fig:logprob_mean_std} illustrates that: the mean log-probability for the most probable adjective (`introverted') is -5.7, whereas it drops to -20.0 for the least probable adjective (`unenvious'). 
Despite this difference, variances in log-probabilities between stories for a specific trait adjective convey meaningful signals. 
For example, the probability of using `unenvious' for trait description is generally low but it is even lower for some stories while relatively higher for others. 
We plot the correlation between log-probabilities for different adjectives (Figure \ref{fig:corr_matrix}) to analyze the variance.
The correlation matrix demonstrates the consistency of log-probabilities measured by LLMs.
In more detail, we observe that a strong positive correlation among log-probabilities of trait adjectives shares the same personality trait and pole, 
and a strong negative correlation among those of opposite poles. 
This evidence supports our assertion that log-probability differences effectively capture the accuracy of an adjective being a description of the person.


\subsection{Trait Descriptive Adjectives}
\label{asec:sub:tda}
There are multiple sets of trait descriptive adjectives in a literature \cite{goldberg1992development, goldberg2013alternative, saucier1997effects} with varying numbers of adjectives and different ways of presentation (e.g. unipolar v.s. bipolar).
We adopt a widely-used trait adjective sets, the 100 unipolar adjectives set \cite{goldberg1992development}. 
100 trait adjectives are listed in Table \ref{tab:tda_100_dict}. 

\newgeometry{margin=0.5in}

\begin{table}[h!]
    \caption{100 unipolar trait descriptive adjectives grouped by Big Five traits and the pole.
     Pole refers to one end of a personality trait's continuum. Each of the five traits is considered a spectrum between two opposite extremes, or poles.
    }
    \vspace{5pt}
    \centering
    \begin{tabular}{
        >{\centering\arraybackslash}p{2.2cm}
        >{\centering\arraybackslash}p{2.2cm}
        >{\centering\arraybackslash}p{2.2cm}
        >{\centering\arraybackslash}p{2.2cm}
        >{\centering\arraybackslash}p{2.2cm}
        >{\centering\arraybackslash}p{2.2cm}
        >{\centering\arraybackslash}p{2.2cm}
    }
        \toprule
        \textbf{Trait} & \multicolumn{2}{c}{\textbf{Extraversion}} & \multicolumn{2}{c}{\textbf{Neuroticism}} & \multicolumn{2}{c}{\textbf{Agreeableness}} \\
        \midrule
        \textbf{Pole} & (+) & (-) & (+) & (-) & (+) & (-) \\
        \midrule
        \textbf{Adjectives} & extraverted & introverted & anxious & unenvious & kind & cold \\
        & talkative & shy & moody & unemotional & cooperative & unkind \\
        & assertive & quiet & temperamental & relaxed & sympathetic & unsympathetic \\
        & verbal & reserved & envious & imperturbable & warm & distrustful \\
        & energetic & untalkative & emotional & unexcitable & trustful & harsh \\
        & bold & inhibited & irritable & undemanding & considerate & demanding \\
        & active & withdrawn & fretful & & pleasant & rude \\
        & daring & timid & jealous & & agreeable & selfish \\
        & vigorous & bashful & touchy & & helpful & uncooperative \\
        & unrestrained & unadventurous & nervous & & generous & uncharitable \\
        & & & insecure & & & \\
        & & & fearful & & & \\
        & & & self-pitying & & & \\
        & & & high-strung & & & \\
        \bottomrule
    \end{tabular}
    \label{tab:tda_100_dict}
\end{table}

\begin{table}[h!]
    \centering
    \begin{tabular}{
        >{\centering\arraybackslash}p{2.2cm}
        >{\centering\arraybackslash}p{2.2cm}
        >{\centering\arraybackslash}p{2.2cm}
        >{\centering\arraybackslash}p{2.2cm}
        >{\centering\arraybackslash}p{2.2cm}
        >{\centering\arraybackslash}p{2.2cm}
        >{\centering\arraybackslash}p{2.2cm}
    }
        \toprule
        \textbf{Trait} & \multicolumn{2}{c}{\textbf{Conscientiousness}} & \multicolumn{2}{c}{\textbf{Openness}} & \multicolumn{2}{c}{} \\
        \midrule
        \textbf{Pole} & (+) & (-) & (+) & (-) & & \\
        \midrule
        \textbf{Adjectives} & organized & disorganized & intellectual & unintellectual & & \\
        & systematic & careless & creative & unintelligent & & \\
        & thorough & unsystematic & complex & unimaginative & & \\
        & practical & inefficient & imaginative & uncreative & & \\
        & neat & undependable & bright & simple & & \\
        & efficient & impractical & philosophical & unsophisticated & & \\
        & careful & negligent & artistic & unreflective & & \\
        & steady & inconsistent & deep & imperceptive & & \\
        & conscientious & haphazard & innovative & uninquisitive & & \\
        & prompt & sloppy & introspective & shallow & & \\
        \bottomrule
    \end{tabular}
\end{table}

\restoregeometry

\subsection{Tokenization of Adjectives}
\label{asec:sub:tokenization}
As LLMs model next-token responses, a log-probability of an adjective is a sum of log-probabilities of tokens that consist the adjective.
A single adjective is tokenized into multiple tokens. For example, an adjective `sophisticated' is tokenized into four tokens with \texttt{Llama3} tokenizer: `s', `oph', `istic', and `ated'. Therefore, to compute the log-probability for `sophisticated' with the prompt (Figure \ref{box:question_example_main}), we compute four instances of log-probability: (1) log-probability of `s' given the prompt, (2) log-probability of `oph' given the prompt appended with `s', (3) log-probability of `istic' given the the prompt appended with `soph', and (4) log-probability of `ated' given the prompt appended with `sophistic'. Four instances of log-probability are added to measure the log-probability for `sophisticated'. Similar process is repeated for each trait adjectives.

\subsection{Decoding Parameter: Temperature}
\label{asec:sub:temp_sweep_method}
Result in the main text is obtained with sampling temperature $T = 1.0$. We can compute the log-probability at different sampling temperatures with a single round of computation at $T = 1.0$, if we have access to log-probabilities for a complete set of vocabularies. This can be done with an efficient LLM serving system \cite{vllm}. The process below describes the computation process at a token level.

Let $N$ be a vocabulary size of a language model, $W$ a list of conditioning tokens, and $w_{\text{target}}$ a next token of which we want to compute log-probability.
Our goal is to compute the log-probability $\text{LP}(w_{\text{target}} | W; T_{\text{t}})$ at a target temperature $T_{\text{t}}$ given a full access to log-probabilities at an original temperature $T_{\text{o}}$, i.e. we know $\text{LP}(w_i | W ; T_{\text{o}})$ for $i \in [1,N]$. We can use the following relation

\vspace{-15pt}
\begin{center}
\[
\text{LP}(w_{\text{target}} | W; T_{\text{t}}) = \log\left(\frac{
\exp\left(\frac{l_{\text{target}}}{T_{\text{t}}}\right) \
}{
\sum\limits_{i=1}^{N} \exp\left(\frac{l_{i}}{T_{\text{t}}}\right)
}\right) = \log\left(\frac{1}{
\sum\limits_{i=1}^{N} \exp\left(\frac{l_i-l_{\text{target}}}{T_{\text{t}}}\right)
}
\right)
\]
\[
= \log\left(\frac{1}{
\sum\limits_{i=1}^{N} \exp\left(\frac{l_i-l_{\text{target}}}{T_{\text{o}}}\cdot\frac{T_{\text{o}}}{T_{\text{t}}}\right)
}
\right) = \log\left(
\frac{1}{
\sum\limits_{i=1}^{N} \exp\left(\left(
\text{LP}(w_i | W; T_{\text{o}}) - \text{LP}(w_{\text{target}} | W; T_{\text{o}})
\right)\cdot\frac{T_{\text{o}}}{T_{\text{t}}}
\right)
}
\right)
\]
\end{center}

where $l_i$ represents a logit for the $i$-th token. Computation at an adjective level is straightforward by using token-level computations repeatedly for each token consisting an adjective (\ref{asec:sub:tokenization}).

\section{Dataset}
Synthetic stories are adopted from the \DATASET dataset~\cite{personallm}. These stories were generated using \texttt{GPT-4-0613} by prompting with prescribed Big-Five personality traits (e.g., "You are a character who is extroverted, agreeable, conscientious, emotionally stable, and closed to experience.") along with an instruction "Please share a personal story in 800 words. Do not explicitly mention your personality traits in the story." The authors generated 10 stories for each of 32 combinations of binary Big-Five personality traits, and filtered out stories that explicitly contained trait-related lexicons. This process resulted in a total of 208 synthetic stories. We use these 208 stories as an input for log-probability measurement, and employ their binary Big-Five personality labels for the prediction accuracy measurement.
\label{asec:dataset}
\section{Additional Experiment Results}
\label{asec:additional}

\subsection{Singular Values of SVD}
\label{asec:sub:svd_sing_val}
\begin{figure*}[h]
    \centering
    \captionsetup{font=small}
    \includegraphics[width=0.60\linewidth]{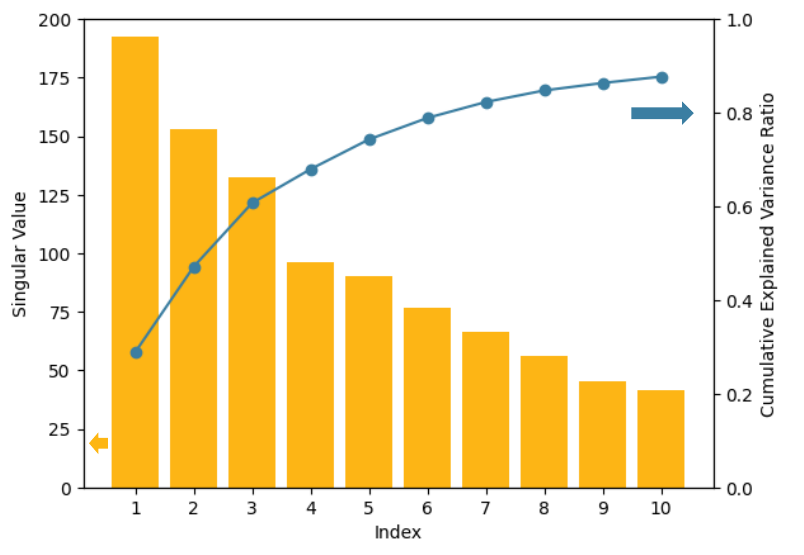}
    \vspace{-5pt}
    \caption{
    Singular values corresponding to each principal component (gold) and cumulative explained variance ratio (blue). At the fifth principal component, cumulative explained variance ratio is 0.743.
    }
    \label{fig:exp_ratio}
    \vspace{-10pt}
\end{figure*}

We present the statistics of the singular values in Figure~\ref{fig:exp_ratio}. 
The gold-colored bars represent the singular values sorted by magnitude, and the blue line illustrates the cumulative explained variance ratio. 
The explained variance ratio measures the proportion of the total variance in the original dataset that is accounted for by each principal component; it is calculated as the ratio of a principal component's eigenvalue to the sum of the eigenvalues of all principal components. 
We observe a significant drop between the fifth and sixth singular values and the cumulative explained variance ratio at the fifth principal component is 0.743. 
This fact indicates that top-5 components sufficiently explain the variance of the latent dimensions.

\subsection{Analyzing Elements of Singular Vectors}
\label{asec:sub:factor_loading}
In the SVD $\tilde{X} = U\Sigma V^\top$, the relation between the \textit{i}-th original dimension and the \textit{j}-th principal component is quantified by the matrix element $V_{ij}$.
In terms of our approach, it is equivalent to the explainability of likelihood of the \textit{i}-th trait adjective (e.g., `introverted') by the \textit{j}-th factor (e.g., extraversion).
Therefore, by analyzing trait adjectives that have the largest or smallest elements in one principal component, we can infer the alignment of that principal component with a Big Five personality trait. 
For example, if adjectives `assertive' and `talkative' have the largest elements while `shy' and `introverted' have the smallest elements in the first principal component, the first principal component is aligned with the extraversion factor.

Table \ref{tab:top_10_bottom_10} shows 10 adjectives with the largest elements and 10 adjectives with the smallest elements for each principal component. Although each dimension does not have a perfect correspondence with a specific Big Five personality trait, we are able to observe tendency. For example, the adjectives having the largest and smallest elements in the first principal component are mostly related with extraversion factor, including `energetic' (0.232), `daring' (0.180), and `quiet' (-0.216). Repeating the analysis for other principal components, we report that the dimensions 1 through 5 are related with Big Five extraversion, openness, agreeableness, neuroticism, and conscientiousness factors, respectively.

We further investigate the relation of Big Five traits with principal components by examining the sign of elements in Table \ref{tab:combined_extraversion_openness}. From our previous observation in Table \ref{tab:top_10_bottom_10}, we focus on the signs of elements of extraversion, openness, agreeableness, neuroticism, and conscientiousness adjectives in principal components 1 through 5, respectively.
Adjectives of the same pole have same sign with few exceptions.
For example, in the first principal component, adjectives associated with the (+) pole of extraversion (`energetic', \dots, `verbal') have positive elements, while adjectives associated with the (-) pole of extraversion (`unadventurous', \dots, `quiet') have negative elements.
Based on these findings we claim that principal components from SVD and Big Five personality traits have a one-to-one correspondence.

\newgeometry{margin=0.5in}

\begin{table}[h]
\centering

\caption{10 adjectives with the largest and smallest loadings in each principal component, from dimension 1 to 5. Extraversion adjectives predominantly have extreme elements in the first principal component, agreeableness adjectives for dimension 3, neuroticism adjectives for dimension 4, and conscientious adjectives for dimension 5. This tendency reveals that each principal component corresponds to a specific Big Five trait.
}

\vspace{10pt}

\small
\resizebox{0.6\linewidth}{!}{%
\begin{tabular}{cccc|cccc}
\toprule
\multicolumn{4}{c|}{\textbf{Dimension 1}} & \multicolumn{4}{c}{\textbf{Dimension 2}} \\
\midrule
Adjective & Factor & Pole & Loading & Adjective & Factor & Pole & Loading \\
\midrule
careless & CON & - & 0.268 & anxious & NEU & + & 0.202 \\
energetic & EXT & + & 0.232 & introverted & EXT & - & 0.190 \\
disorganized & CON & - & 0.200 & withdrawn & EXT & - & 0.187 \\
haphazard & CON & - & 0.188 & distrustful & AGR & - & 0.187 \\
daring & EXT & + & 0.180 & inhibited & EXT & - & 0.180 \\
vigorous & EXT & + & 0.176 & reserved & EXT & - & 0.176 \\
extraverted & EXT & + & 0.174 & insecure & NEU & + & 0.176 \\
unrestrained & EXT & + & 0.170 & cold & AGR & - & 0.172 \\
high-strung & NEU & + & 0.168 & unemotional & NEU & - & 0.167 \\
unsystematic & CON & - & 0.156 & irritable & NEU & + & 0.159 \\
\midrule
intellectual & OPN & + & -0.081 & active & EXT & + & -0.046 \\
steady & CON & + & -0.087 & innovative & OPN & + & -0.049 \\
bashful & EXT & - & -0.092 & generous & AGR & + & -0.054 \\
untalkative & EXT & - & -0.106 & vigorous & EXT & + & -0.067 \\
withdrawn & EXT & - & -0.137 & artistic & OPN & + & -0.075 \\
shy & EXT & - & -0.171 & bold & EXT & + & -0.081 \\
introverted & EXT & - & -0.188 & imaginative & OPN & + & -0.087 \\
reserved & EXT & - & -0.200 & creative & OPN & + & -0.114 \\
quiet & EXT & - & -0.216 & energetic & EXT & + & -0.115 \\
introspective & OPN & + & -0.238 & daring & EXT & + & -0.147 \\
\bottomrule
\end{tabular}
}
\resizebox{1.0\linewidth}{!}{%
\begin{tabular}{cccc|cccc|cccc}
\toprule
\multicolumn{4}{c|}{\textbf{Dimension 3}} & \multicolumn{4}{c|}{\textbf{Dimension 4}} & \multicolumn{4}{c}{\textbf{Dimension 5}} \\
\midrule
Adjective & Factor & Pole & Loading & Adjective & Factor & Pole & Loading & Adjective & Factor & Pole & Loading \\
\midrule
kind & AGR & + & 0.309 & nervous & NEU & + & 0.304 & haphazard & CON & - & 0.257 \\
helpful & AGR & + & 0.267 & anxious & NEU & + & 0.272 & shy & EXT & - & 0.173 \\
generous & AGR & + & 0.262 & high-strung & NEU & + & 0.251 & disorganized & CON & - & 0.167 \\
considerate & AGR & + & 0.251 & complex & OPN & + & 0.233 & relaxed & NEU & - & 0.162 \\
warm & AGR & + & 0.232 & creative & OPN & + & 0.192 & sloppy & CON & - & 0.162 \\
steady & CON & + & 0.190 & fretful & NEU & + & 0.167 & careless & CON & - & 0.150 \\
organized & CON & + & 0.189 & fearful & NEU & + & 0.167 & introverted & EXT & - & 0.147 \\
sympathetic & AGR & + & 0.189 & imaginative & OPN & + & 0.158 & inconsistent & CON & - & 0.144 \\
pleasant & AGR & + & 0.180 & organized & CON & + & 0.154 & unadventurous & EXT & - & 0.144 \\
neat & CON & + & 0.173 & daring & EXT & + & 0.149 & bashful & EXT & - & 0.143 \\
\midrule
introverted & EXT & - & -0.080 & unexcitable & NEU & - & -0.095 & conscientious & CON & + & -0.128 \\
harsh & AGR & - & -0.081 & uncooperative & AGR & - & -0.100 & cold & AGR & - & -0.130 \\
insecure & NEU & + & -0.084 & uninquisitive & OPN & - & -0.109 & thorough & CON & + & -0.138 \\
cold & AGR & - & -0.090 & selfish & AGR & - & -0.126 & organized & CON & + & -0.167 \\
moody & NEU & + & -0.091 & unsophisticated & OPN & - & -0.132 & systematic & CON & + & -0.188 \\
irritable & NEU & + & -0.099 & unkind & AGR & - & -0.135 & demanding & AGR & - & -0.199 \\
withdrawn & EXT & - & -0.102 & uncharitable & AGR & - & -0.139 & unemotional & NEU & - & -0.204 \\
complex & OPN & + & -0.104 & undemanding & NEU & - & -0.145 & harsh & AGR & - & -0.204 \\
distrustful & AGR & - & -0.157 & unsympathetic & AGR & - & -0.162 & efficient & CON & + & -0.225 \\
rude & AGR & - & -0.165 & rude & AGR & - & -0.169 & assertive & EXT & + & -0.230 \\
\bottomrule
\end{tabular}
}
\label{tab:top_10_bottom_10}
\end{table}

\restoregeometry

\newgeometry{margin=0.5in}

\begin{table*}[h]
\centering

\caption{Elements (loadings) of extraversion, openness, agreeableness, neuroticism, and conscientiousness adjectives for principal components 1 to 5, respectively.
Adjectives belonging to the same pole (either (+) or (-)) maintain same polarity, except for few exceptions:
`neat' in conscientiousness trait, whose element is close to 0;
`complex', `introspective', `deep', `intellectual' in openness trait.
Combining with the results from Table \ref{tab:top_10_bottom_10}, we conclude that latent factors from SVD resemble the Big Five factors observed in human personality assessment \cite{goldberg1992development}.
}

\small
\resizebox{1.0\linewidth}{!}{%
\begin{tabular}{ccc|ccc|ccc|ccc}
\toprule
\multicolumn{6}{c|}{\textbf{Extraversion Adjectives in Dimension 1}} & \multicolumn{6}{c}{\textbf{Openness Adjectives in Dimension 2}} \\
\midrule
\multicolumn{3}{c|}{\textbf{Positive Loadings}} & \multicolumn{3}{c|}{\textbf{Negative Loadings}} & \multicolumn{3}{c|}{\textbf{Positive Loadings}} & \multicolumn{3}{c}{\textbf{Negative Loadings}} \\
\midrule
Adjective & Pole & Loading & Adjective & Pole & Loading & Adjective & Pole & Loading & Adjective & Pole & Loading \\
\midrule
energetic  & + & 0.233 & unadventurous & - & -0.010 & uncreative      & - & 0.099  & philosophical  & + & 0.000 \\
daring  & + & 0.181 & inhibited & - & -0.041 & uninquisitive   & - & 0.098  & bright         & + & -0.028 \\
vigorous  & + & 0.177 & timid & - & -0.077 & unintelligent   & - & 0.095  & innovative     & + & -0.049 \\
extraverted  & + & 0.175 & bashful & - & -0.092 & unintellectual  & - & 0.094  & artistic       & + & -0.075 \\
unrestrained  & + & 0.171 & untalkative & - & -0.106 & imperceptive    & - & 0.091  & imaginative    & + & -0.087 \\
bold  & + & 0.154 & withdrawn & - & -0.137 & simple          & - & 0.090  & creative       & + & -0.114 \\
assertive  & + & 0.117 & shy & - & -0.171 & shallow         & - & 0.087  &                &   &        \\
talkative  & + & 0.117 & introverted & - & -0.188 & unreflective    & - & 0.083  &                &   &        \\
active  & + & 0.102 & reserved & - & -0.200 & unimaginative   & - & 0.082  &                &   &        \\
verbal  & + & 0.088 & quiet & - & -0.216 & complex         & + & 0.071  &                &   &        \\
         &   &       &       &   &        & introspective   & + & 0.063  &                &   &        \\
         &   &       &       &   &        & unsophisticated & - & 0.027  &                &   &        \\
         &   &       &       &   &        & deep            & + & 0.019  &                &   &        \\
         &   &       &       &   &        & intellectual    & + & 0.011 &                &   &         \\
\bottomrule
\end{tabular}
}
\label{tab:combined_extraversion_openness}
\end{table*}

\begin{table*}[h!]
\centering
\small
\resizebox{1.0\linewidth}{!}{%
\begin{tabular}{ccc|ccc|ccc|ccc}
\toprule
\multicolumn{6}{c|}{\textbf{Agreeableness Adjectives in Dimension 3}} & \multicolumn{6}{c}{\textbf{Neuroticism Adjectives in Dimension 4}} \\
\midrule
\multicolumn{3}{c|}{\textbf{Positive Loadings}} & \multicolumn{3}{c|}{\textbf{Negative Loadings}} & \multicolumn{3}{c|}{\textbf{Positive Loadings}} & \multicolumn{3}{c}{\textbf{Negative Loadings}} \\
\midrule
Adjective & Pole & Loading & Adjective & Pole & Loading & Adjective & Pole & Loading & Adjective & Pole & Loading \\
\midrule
kind         & + & 0.309 & uncharitable   & - & -0.020 & nervous        & + & 0.304 & unenvious      & - & -0.045 \\
helpful      & + & 0.267 & demanding      & - & -0.034 & anxious        & + & 0.272 & imperturbable  & - & -0.059 \\
generous     & + & 0.262 & unsympathetic  & - & -0.047 & high-strung    & + & 0.251 & relaxed        & - & -0.067 \\
considerate  & + & 0.251 & selfish        & - & -0.049 & fretful        & + & 0.167 & unemotional    & - & -0.079 \\
warm         & + & 0.232 & unkind         & - & -0.058 & fearful        & + & 0.167 & unexcitable    & - & -0.095 \\
sympathetic  & + & 0.189 & uncooperative  & - & -0.063 & insecure       & + & 0.119 & undemanding    & - & -0.145 \\
pleasant     & + & 0.180 & harsh          & - & -0.081 & moody          & + & 0.073 &                &   &        \\
agreeable    & + & 0.149 & cold           & - & -0.090 & emotional      & + & 0.067 &                &   &        \\
cooperative  & + & 0.141 & distrustful    & - & -0.157 & temperamental  & + & 0.061 &                &   &        \\
trustful     & + & 0.087 & rude           & - & -0.165 & envious        & + & 0.045 &                &   &        \\
             &   &       &                &   &        & irritable      & + & 0.042 &                &   &        \\
             &   &       &                &   &        & touchy         & + & 0.029 &                &   &        \\
             &   &       &                &   &        & jealous        & + & 0.027 &                &   &        \\
             &   &       &                &   &        & self-pitying   & + & 0.011 &                &   &        \\
\bottomrule
\end{tabular}
}
\label{tab:combined_agreeableness_neuroticism}
\end{table*}

\begin{table*}[h!]
\centering
\small
\begin{tabular}{ccc|ccc}
\toprule
\multicolumn{6}{c}{\textbf{Conscientiousness Adjectives in Dimension 5}} \\
\midrule
\multicolumn{3}{c|}{\textbf{Positive Loadings}} & \multicolumn{3}{c}{\textbf{Negative Loadings}} \\
\midrule
Adjective & Pole & Loading & Adjective & Pole & Loading \\
\midrule
haphazard      & - & 0.257 & prompt         & + & -0.022 \\
disorganized   & - & 0.167 & careful        & + & -0.056 \\
sloppy         & - & 0.162 & practical      & + & -0.102 \\
careless       & - & 0.150 & steady         & + & -0.122 \\
inconsistent   & - & 0.144 & conscientious  & + & -0.128 \\
impractical    & - & 0.123 & thorough       & + & -0.138 \\
unsystematic   & - & 0.121 & organized      & + & -0.167 \\
inefficient    & - & 0.101 & systematic     & + & -0.188 \\
undependable   & - & 0.072 & efficient      & + & -0.225 \\
negligent      & - & 0.044 &                &   &        \\
neat           & + & 0.003 &                &   &        \\
\bottomrule
\end{tabular}
\label{tab:conscientiousness_con}
\end{table*}

\restoregeometry

\subsection{Personality Prediction with SVD}
\label{asec:sub:svd_prediction_procedure}
In this section, we elaborate how we can predict the personality traits based on SVD analysis. 
We compare the sign of a column of the factor matrix $U \in \mathbb{R}^{N \times k}$ with Big Five personality labels $L \in \{0, 1\}^{N \times 5}$, where $k=5$ in this work.
However, since we have not yet determined which principal component corresponds to each personality trait, we need a method to establish this correspondence.
To achieve this, we construct the accuracy matrix, $P$, where each element $P_{ij}$ is the prediction accuracy of $j$-th personality trait with $i$-th principal component.

$P$ is in Table \ref{tab:ours_accuracy_appendix}. An accuracy of 1 is close to a perfect prediction, while an accuracy of 0.5 is close to a random guess. When comparing the first column of $U$ with the Big Five extraversion label, the accuracy is 0.899. Comparing the first column of $U$ with Big Five labels other than extraversion gives accuracy of 0.505, 0.581, 0.591, 0.524, all close to a random guess.
Notably, for each principal component, there exist only \emph{one} Big Five label that yields high prediction accuracy. Columns 1 to 5 have high accuracy when compared with extraversion, openness, agreeableness, neuroticism, conscientiousness labels, respectively. This is the same one-to-one correspondence relation obtained in \ref{asec:sub:factor_loading}.

\begin{table}[h]
    \small
    \captionsetup{font=small}
    \centering
    \caption{Accuracy matrix \( P \) comparing the signs of the first five principal components (rows) with the Big Five personality traits (columns). Each entry \( P_{ij} \) represents the accuracy of predicting the \( j \)-th personality trait from the \( i \)-th principal component. The matrix is derived from the training dataset using SVD with \( k=5 \). The highest accuracy in each row is highlighted in bold, illustrating a one-to-one correspondence between principal components and personality traits.}
    \label{tab:ours_accuracy_appendix}
    \vspace{10pt}
    \begin{tabular}{c|c|c|c|c|c}
    \toprule
    \multirow{1}{*}{Index} & 
    \multirow{1}{*}{Extraversion} &
    \multirow{1}{*}{Agreeableness} &
    \multirow{1}{*}{Conscientiousness} &
    \multirow{1}{*}{Neuroticism} &
    \multirow{1}{*}{Openness} \\
     \midrule
    1 & \textbf{0.899} & 0.581 & 0.591 & 0.505 & 0.524 \\\midrule
    2 & 0.663 & 0.586 & 0.510 & 0.548 & \textbf{0.798} \\\midrule
    3 & 0.548 & \textbf{0.817} & 0.596 & 0.615 & 0.567 \\\midrule
    4 & 0.533 & 0.514 & 0.562 & \textbf{0.726} & 0.600 \\\midrule
    5 & 0.524 & 0.572 & \textbf{0.803} & 0.543 & 0.582 \\
    \bottomrule
    \end{tabular}
\end{table}
\section{Baseline Experiment Details}
\label{asec:baseline}

\subsection{Fine-tuning Encoder Models}
\label{asec:sub:baseline_encoder}
As a baseline, we fine-tune the DeBERTa-V3 Large model~\cite{he2021debertav3} using \DATASET dataset. 
We place a 5-way classifier on top of the output of the [CLS] token from the encoder model. 
We utilize binary cross-entropy loss for fine-tuning.
The AdamW optimizer~\cite{loshchilov2017fixing} is utilized along with a cosine learning rate scheduler and a linear warm-up. 
The number of warm-up steps is set to 300. 
We divide \DATASET dataset into 80\% for training, 10\% for validation, and 10\% for test.
Hyperparameter tuning is performed on the validation set and the test set is evaluated only once using the best-performing hyperparameter configuration from the validation set.
We perform a hyperparameter sweep over batch size, learning rate, weight decay, and number of training epochs. 
The hyperparameter configurations we use are: $\{4, 8, 16, 32\}$ for batch size, $\{5e-5, 1e-4, 2e-4, 4e-4\}$ for learning rate, $\{0.01, 0.02\}$ for weight decay, and $\{5, 10, 15, 20\}$ for the number of training epochs.

\subsection{Prompting LLMs to Evaluate Personality}
\label{asec:sub:baseline_prompting}
Prior works prompt language models to assess the Big Five personality score in a Likert scale, usually on a spectrum of 1 to 5. Various prompts have been designed: to list a few, `please rate how accurately this describes you on a scale from 1 to 5 (where 1 = "very inaccurate", 2 = "moderately inaccurate", 3 = "neither accurate nor inaccurate", 4 = "moderately accurate", and 5 = "very accurate")' \cite{serapio2023personality} `Rate the personality of the person called "user" on the Big Five personality dimensions. Pay attention to how people’s personalities might be reflected in the way they respond to questions and what they share about themselves. Provide your response on a scale from 1 to 5 for the traits Openness, Conscientiousness, Extraversion, Agreeableness, and Neuroticism. Provide only the numbers.' \cite{peters2024large}, and `What is your guess for the big-five personality traits of someone who said "{\textit{text}}", answer low or high with bullet points for the five traits? It does not have to be fully correct. You do not need to explain the traits. Do not show any warning after.' \cite{amin2023will}. Authors of \DATASET dataset \cite{personallm} also utilized the prompting method. The published accuracy value from the paper is taken as a baseline method accuracy.

\end{document}